\documentclass[conference]{IEEEtran}
\IEEEoverridecommandlockouts

\usepackage{balance}
\usepackage{cite}
\usepackage{float}
\usepackage{amsmath,amssymb,amsfonts}
\usepackage{algorithmic}
\usepackage{graphicx}
\usepackage{textcomp}
\usepackage{xcolor}
\usepackage{acronym}
\usepackage{tablefootnote}
\def\BibTeX{{\rm B\kern-.05em{\sc i\kern-.025em b}\kern-.08em
    T\kern-.1667em\lower.7ex\hbox{E}\kern-.125emX}}

\newcommand{\figref}[1]{Fig. \ref{#1}}
\newcommand{\secref}[1]{Section \ref{#1}}
\newcommand{\tabref}[1]{Tab. \ref{#1}}
\renewcommand{\quote}[1]{``#1''}

\def\anonymize{0}

\if\anonymize1
	
	\else
	
\fi

\begin{document}
\acrodef{snn}[SNN]{Spiking Neural Network}
\acrodef{ann}[ANN]{Artificial Neural Network}
\acrodef{shd}[SHD]{Spiking Heidelberg Digits}
\acrodef{ssc}[SSC]{Spiking Speech Commands}
\acrodef{sota}[SOTA]{state of the art}
\acrodef{lif}[LIF]{Leaky-Integrate and Fire}
\acrodef{lod}[LOD]{Leading-One Detector}
\acrodef{lopd}[LOPD]{Leading-One Position Detector}

\title{Spiking Neural Network Accelerator Architecture for Differential-Time Representation using Learned Encoding}


\author{\IEEEauthorblockN{Daniel Windhager\IEEEauthorrefmark{1}, Lothar Ratschbacher\IEEEauthorrefmark{2}, Bernhard A. Moser$^{**}$\IEEEauthorrefmark{3}\thanks{$^{**}$double affiliation: Software Competence Center Hagenberg (SCCH), 4232 Hagenberg, Austria\newline\newline This work was supported (1) by Silicon Austria Labs (SAL), owned by the Republic of Austria, the Styrian Business Promotion Agency (SFG), the federal state of Carinthia, the Upper Austrian Research (UAR), and the Austrian Association for the Electric and Electronics Industry (FEEI), (2) by the ``University SAL Labs'' initiative of Silicon Austria Labs (SAL) and its Austrian partner universities for applied fundamental research for electronic based systems, (3) by Austrian ministries BMK, BMDW, and the State of Upper-Austria in in the frame of SCCH and its project S3AI, part of the COMET Programme managed by FFG, and (4) by the {\it NeuroSoC} project funded under the Horizon Europe Grant Agreement number 101070634.}, Michael Lunglmayr\IEEEauthorrefmark{4}}
\IEEEauthorblockA{\IEEEauthorrefmark{1}Intelligent Wireless Systems, Silicon Austria Labs, Linz, Austria\\daniel.windhager@silicon-austria.com}
\IEEEauthorblockA{\IEEEauthorrefmark{2}Intelligent Wireless Systems, Silicon Austria Labs, Linz, Austria\\lothar.ratschbacher@silicon-austria.com}
\IEEEauthorblockA{\IEEEauthorrefmark{3}Institute of Signal Processing, Johannes Kepler University, JKU SAL IWS Lab, Linz, Austria\\bernhard.moser@jku.at}
\IEEEauthorblockA{\IEEEauthorrefmark{4}Institute of Signal Processing, Johannes Kepler University, Linz, Austria\\michael.lunglmayr@jku.at}}

\maketitle

\begin{abstract}
\acp{snn} have garnered attention over recent years due to their increased energy efficiency and advantages in terms of operational complexity compared to traditional \acp{ann}. Two important questions when implementing \acp{snn} are how to best encode existing data into spike trains and how to efficiently process these spike trains in hardware. This paper addresses both of these problems by incorporating the encoding into the learning process, thus allowing the network to learn the spike encoding alongside the weights. Furthermore, this paper proposes a hardware architecture based on a recently introduced differential-time representation for spike trains allowing decoupling of spike time and processing time. Together these contributions lead to a feedforward \ac{snn} using only \ac{lif} neurons that surpasses 99\% accuracy on the MNIST dataset while still being implementable on medium-sized FPGAs with inference times of less than 295$\mu{}$s.
\end{abstract}


\section{Introduction}
\acp{snn} have shown promising results on several datasets ranging from classical image processing tasks such as MNIST \cite{mnist}, to audio processing tasks such as on the \ac{shd} dataset\cite{shd_dataset} and the \ac{ssc} dataset by Tensorflow\cite{tf_speechcommands}. In order to efficiently process \acp{snn}, accelerator architectures are needed, of which multiple have been proposed in literature\cite{darwin,energy_efficient_neuromorph_arch,loihi,noc,snava,TCASSPIKEMNIST21,truenorth,low_power_fpga_snn_acc,spinnaker}. A previously presented architecture\cite{aicas2024} introduced the concept of differential-time for efficiently representing spikes in hardware, by only storing the time difference between spikes, which is also used in this work as detailed in \secref{sec:hw_arch}. However, instead of using rotation registers to process the spikes as in \cite{aicas2024}, an efficient spike sorting architecture is proposed, allowing to significantly speed up processing while reducing resource requirements. Furthermore, instead of handling the spike time locally, i.e. separately for each spike, this architecture processes spike time globally for the entire layer, further simplifying the design. Regardless of the accelerator being used, the network expects spike trains as inputs which therefore necessitates encoding the input data. A second important differentiating feature of this work compared to \cite{aicas2024} is that the \ac{snn} architecture presented in this paper allows the network to learn this encoding, which somewhat resembles the attention mechanism found in modern transformer architectures\cite{transformers}. This enables the network to achieve accuracy results comparable to \ac{sota} \acp{snn} on the MNIST dataset\cite{odesa,spiker,spiker+,minitaur}, while only using \ac{lif} neurons throughout the network, thus allowing for a smaller and faster implementation. \secref{sec:encoding} introduces the concept of incorporating the encoding into the learning process. The performance of the accelerator and the network are shown in \secref{sec:results}.
\subsection{\ac{lif} neuron model}
The \acp{snn} of this work are only composed of \ac{lif} neurons (with reset by subtraction), which follow the basic equation 
\begin{align}
\begin{split}
	P_{k} &= P_{k-1}\beta^{\Delta t} + \sum_i w_i s_i - s_{out}\\
	s_{out} &= \begin{cases}
	\mbox{$\theta$,} & \mbox{if } P \ge \theta\\
	\mbox{0,} & \mbox{if } P < \theta,
	\end{cases}
\end{split}
\label{eq:lif_neuron}
\end{align}
where $P$ is the neuron potential, $\beta$ is the decay rate of the potential, $\Delta t$ is the time that passed since the last event, $k$ is the time where at least one input spike $s_i$ occurred, $\sum_i w_i s_i$ is the weighted sum of the input spikes (at the current event) and their corresponding weights, $s_{out}$ is responsible for resetting the neuron potential after a spike is triggered and $\theta$ is the threshold. Without loss of generality, we set $\theta = 1$ and $\beta=0.5$ in this work (for an in-depth explanation why this does not restrict generality see \cite{aicas2024}). These \ac{lif} neurons are then arranged in multiple layers to form the feedforward \ac{snn}.

\section{Hardware architecture}
\label{sec:hw_arch}

\begin{figure}
	\centering
	\includegraphics[width=0.9\linewidth]{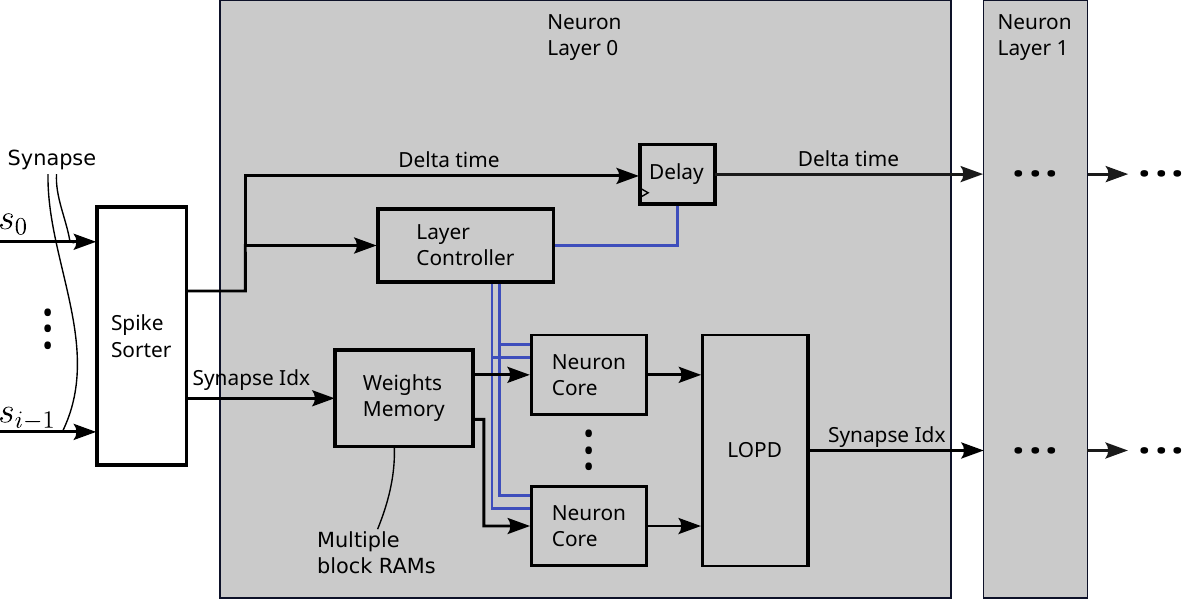}
	\caption{Hardware architecture of network with multiple neuron layers. Data signals are colored black, while control signals are colored blue.}
	\label{fig:hw_arch}
\end{figure}
The entire network consists of multiple fully connected layers, where each layer contains a number of \ac{lif} neurons \eqref{eq:lif_neuron}, each of which only produces positive spikes since this version of the \ac{lif} neuron produced better accuracy results during training (although the architecture also allows using neuron models with negative spikes). The hardware required to accelerate these multi-layer \acp{snn} can be seen in \figref{fig:hw_arch}. The input spike trains are first fed into the Spike Sorter (details below), which is only needed once for the entire network, and is responsible for serializing the spikes by sorting them according to the time of their occurrence. The Spike Sorter outputs the delta time of the next spike as well as the index of the synapse that produced the spike. The delta time is necessary for decaying the neuron potential (see \eqref{eq:lif_neuron}), while the synapse index is used to address the block RAMs to retrieve the weights associated with the sending synapse of the spike. The weights are handed directly to the Neuron Cores, which are responsible for decaying and accumulating the neuron potential as well as generating new spikes. These generated spikes are once again serialized, this time using a \ac{lopd} since sorting is no longer necessary, as the spikes remain sorted throughout the layer. Finally, the decision whether to add the weights or generate new spikes is handled by the Layer Controller. Since the Spike Sorter, Neuron Cores, LOPD, and Layer Controller constitute the main functionality of the entire accelerator, these will be examined in more detail in the next sections.

\subsection{Spike Sorter}
\begin{figure}
	\centering
	\includegraphics[width=0.75\linewidth]{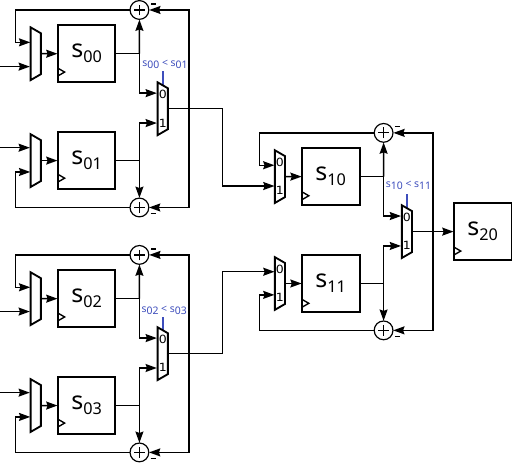}
	\caption{Spike Sorter design with 4 inputs.}
	\label{fig:spike_sorter}
\end{figure}

 The Spike Sorter is responsible for serializing the input spikes that arrive in parallel via the input synapses of the whole network while ensuring that the output spikes are sorted by their occurrence in absolute time. Since each of the spikes is encoded using delta time, simply comparing the delta times between synapses would not work, as their absolute time depends on the previous spikes. Therefore, an efficient way had to be found to allow sorting of the delta time represented spikes across all input synapses. One way to do this would be to accumulate all the delta times for each of the input synapses and subsequently sort them. These absolute times then need to be converted back to delta times, since only the time that passed in between spikes is relevant for decaying the neuron potential. Besides being expensive to implement in hardware (it would diminish the benefits of delta time encoding), it would also lead to overflow since integrators are inherently unstable and will overflow given enough time. A better method to accomplish this, and the method implemented here, is shown in \figref{fig:spike_sorter} using 4 input synapses. The spikes from each synapse are first stored in registers, after which they are compared with their nearest neighbor. The smallest, i.e. earliest, of these spikes is then transferred to the next register in line as seen in \figref{fig:spike_sorter} where $s_{10} = \textit{min}(s_{00}, s_{01})$. After the smaller of the two spikes is propagated, it is subtracted from both previously compared spikes, causing at least one to be set to 0. This is then repeated until all of the spikes have traveled through the structure. This approach not only produces the correct delta time of the next spike but also allows reconstructing the spike index by observing the way of a spike through the Spike Sorter. The index can be seen in the bit vector, combining the comparison bits of a spike. If the upper input to the comparison is chosen, a '0' is set in the bit vector, otherwise a '1' is set, resulting in a bit vector after the last comparison that represents the index of the synapse that produced this spike.

\subsection{Neuron Core}
\begin{figure}
	\centering
	\includegraphics[width=0.65\linewidth]{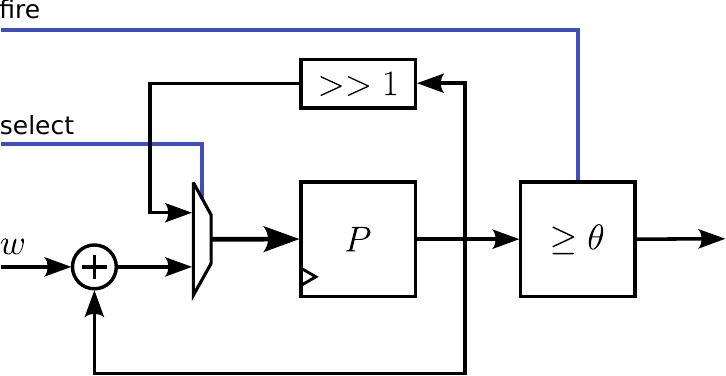}
	\caption{Neuron Core requiring only one register, one comparator, a multiplexer and one adder.}
	\label{fig:neuron_core}
\end{figure}

The output from the Spike Sorter is the delta time since the last spike, as well as the index denoting which synapse the spike originated from. This synapse index is directly used to address the block RAMs as shown in \figref{fig:hw_arch}, which contains the previously learned weights. In order to speed up this operation, multiple parallel block RAMs are combined to enable wide accesses, allowing the hardware to read all necessary weights in a single clock cycle. After the weights are retrieved they are handed to the Neuron Cores, for which the hardware is shown in \figref{fig:neuron_core}. The Neuron Cores are responsible for accumulating weights and decaying the neuron potential, as well as generating new spikes if the threshold $\theta$ is crossed. The \emph{select} control signal determines whether the potential is decayed or the weight $w$ is added to the potential. The decay is realized by executing a single bit shift of the potential, for the number of clock cycles dictated by the delta time of the current spike. Since all Neuron Cores experience the same decay, this can be handled in the Layer Controller and distributed via the \emph{select} signal. After all spikes for a single timestep have been processed, and their respective weights have been added to the neuron potential, the Neuron Core compares the current potential to the threshold $\theta$, and generates a spike if necessary.  

\subsection{\ac{lopd}}
Each of the Neuron Cores can generate a spike, which needs to be propagated to the next layer. The spike inputs of a layer, consisting of a synapse index and delta time respectively, are generated in parallel and thus must be serialized first before processing in the next layer. Finding the indices of all the ones in a bit vector could be trivially accomplished by simply iterating through the vector, however, the time to find all ones scales linearly with the number of bits in the vector. A way to speed up finding the positions is to use a \ac{lod}, or in this case, a \acf{lopd}, for which different architectures and designs have been proposed in the literature, e.g. \cite{fast_lp_lod,vlsi_lod,vlsi_lp_lod,approx_lod}. The \ac{lopd} produces the index of the leading one bit as well as a one-hot representation of the input, which therefore must be run multiple times with the previously detected bit reset, in order to continuously detect all one bits in the vector. Different variants of the \ac{lopd} were synthesized by the authors using Xilinx Vivado (v2023.2), all of which generated the same result on the ZCU102 board, suggesting that Xilinx Vivado recognizes and optimizes for \acp{lopd} specifically, making such a component an ideal choice. 

\subsection{Layer Controller}
The spikes that enter each layer are processed serially. All neurons in a layer experience the same decay, that can thus be generated globally in the layer and distributed to each Neuron Core. Furthermore, the Layer Controller is responsible for ensuring that the thresholding operation, including the accompanying spike generation, is only done once all of the spikes occurring at the same time have been processed. Furthermore, the delta time of the spike being currently processed inside the layer needs to be delayed by the correct amount of cycles until it can be propagated to the next layer (as the delta time is available immediately while the spike requires cycles for processing in the Neuron Cores) since all spikes produced by the Neuron Cores will share the time of the spike that triggered them. 

\section{Learning the encoding}
\label{sec:encoding}
\begin{figure}
	\centering
	\includegraphics[width=0.85\linewidth]{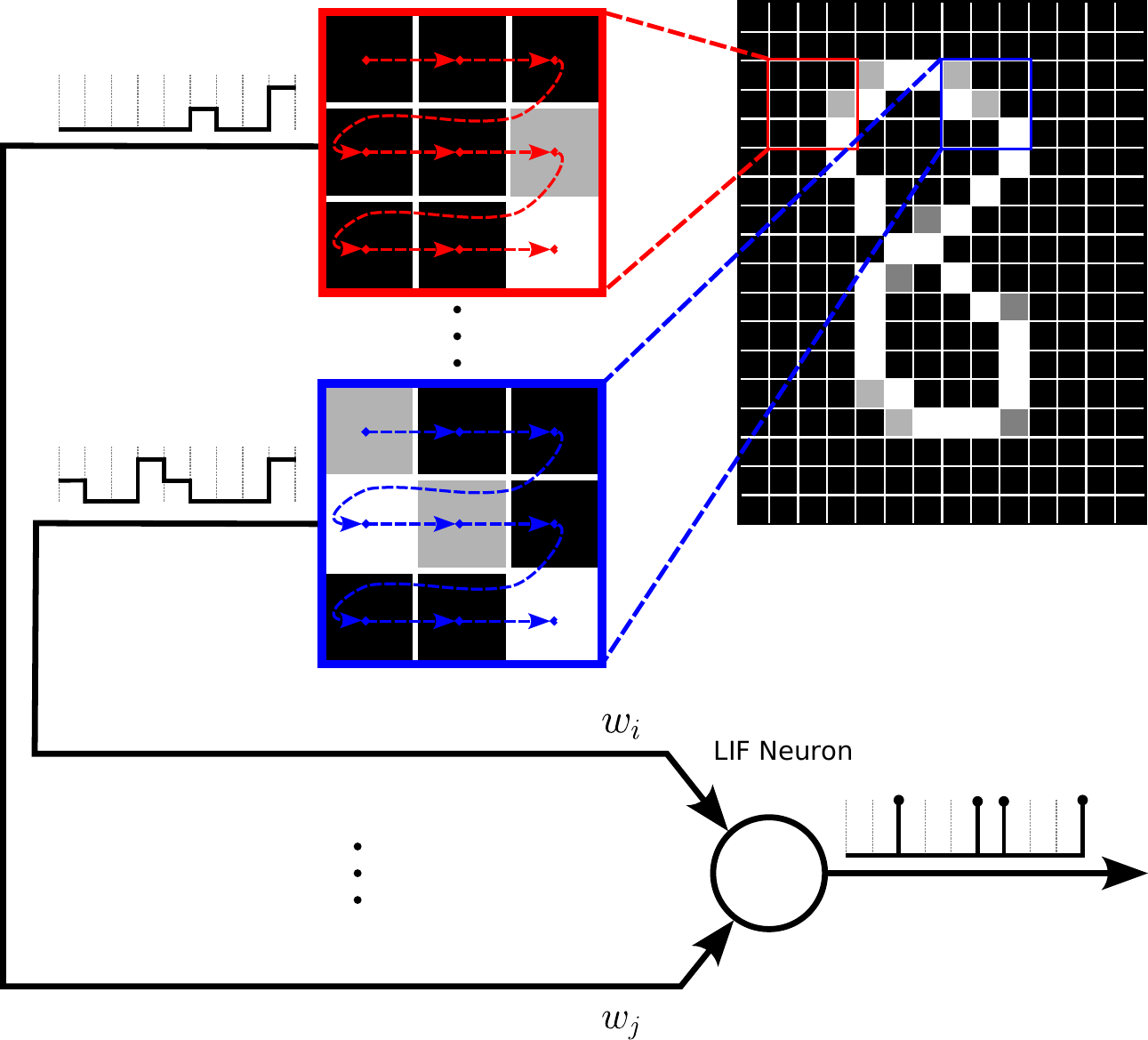}
	\caption{Patch-wise selection of the pixels with subsequent neuron activation with weights learned for encoding.}
	\label{fig:patchwise}
\end{figure}

\begin{figure}
	\centering
	\includegraphics[width=\linewidth]{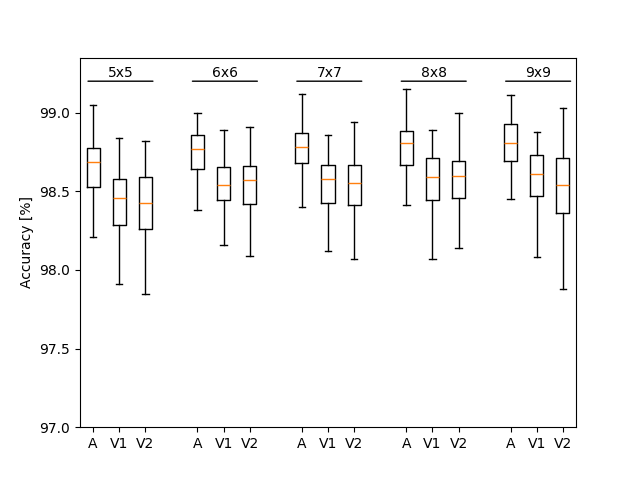}
	\caption{Results of network evaluation on MNIST dataset with patch sizes ranging from 5x5 to 9x9, averaged over 96 runs respectively. Arbitrary weights (A) are quantized to $w=2^i$ (V1) and $w\in\{-1,0,1\}$ (V2).}
	\label{fig:boxplot}
\end{figure}

\ac{snn} neurons expect spike trains as inputs which necessitates a certain spike encoding to generate these spike trains from the original data. This can be achieved using encodings such as rate encoding or latency encoding\cite{snntorch}. We used a different approach, enabling the network to learn an appropriate encoding by feeding pixels as input to \ac{lif} neurons, using a first layer of \ac{lif} neurons for spike encoding. In order for the network to have the ability to learn which input data points, or in this case pixels, are most important, the neurons in the first layer need to be able to access a set of pixels in the image. For the MNIST dataset, this is usually accomplished by having a first layer with 784 ($28\cdot28 = 784$) neurons, each of which receives one previously encoded pixel during multiple timesteps. While used in the past\cite{snntorch}, this approach has several drawbacks, one being a large number of neurons in the first layer. The images in the MNIST dataset are small compared to datasets and images recorded using image sensors typically available today, therefore following the approach of \cite{snntorch} would lead to impractical SNN sizes for many practical applications. As an alternative, we propose to present each neuron in the first layer with a number of pixels, fed in successive order, taken from a certain patch of each image as can be seen in \figref{fig:patchwise}.
Conventionally, this patch of pixels would then be encoded, using for example rate encoding, and then handed to the respective neurons. In this work, the pixels are handed directly to the neuron, without any preprocessing or encoding. Since each of the input spikes to a \ac{lif} neuron, as described in \eqref{eq:lif_neuron} is assigned a weight, after which all the weighted spikes are summed, this would require additional multiplication operations. However, this can be mitigated during the learning process by adjusting the weights in such a way as to avoid multiplication. Two different methods, using quantization-aware training, were tested and evaluated here: the first was to restrict the weights to be of the form $w=2^i$ where $i\in\mathbb{Z}$, which allows using a bit shift instead of multiplication. After learning, this shift is fixed, thus posing only negligible complexity at inference time. For the second method, the weights were restricted to the set $W \in \{-1, 0, 1\}$, which, as most complex operation only requires negation. Each of the two methods was evaluated with 5 different patch sizes $p$$\times$$p$ (ranging from 5$\times$5 to 9$\times$9), with a fixed network structure of Y-800-512-256-10 (where $Y = (28-p+1)^2$ for $p\in[5,9]$). Each of the combinations was trained and evaluated on the MNIST dataset. The results of these evaluations for both restrictions on the weights were repeated and averaged over 96 separate runs. The results can be seen in \figref{fig:boxplot}.
As the figure shows, allowing powers of 2 for the weights did not improve the accuracy significantly compared to the solution which only allowed $+1$, $-1$, or 0. The latter solution actually performs slightly better for some patch sizes, indicating that the network depends more on the presence/absence of certain pixels than on their actual values. This is furthermore supported by the fact that allowing arbitrary weights (shown as A in \figref{fig:boxplot}), i.e. the version that would require multiplications, produced only marginally better accuracy results. 


\section{Simulation and synthesis results}
\label{sec:results}
The accelerator was synthesized for a network having 4 layers of fully connected \ac{lif} neurons, consisting of 800, 512, 256, and 10 neurons respectively. The encoding was done offline using a 9x9 patch size with the weights being restricted to $\{-1, 0, 1\}$. The results from the synthesis using Xilinx Vivado v2023.2 can be seen in \tabref{tab:perf_comparison}. As is evident, the design elaborated in this paper not only achieves comparable inference performance, but also requires fewer resources compared to other \ac{sota} accelerator architectures. Furthermore, the design presented in this paper is \quote{\ac{lif}-only}, i.e. we only use our described variant of \ac{lif} that is multiplier-free, while most other works found in literature that achieve more than 99\% accuracy on the MNIST dataset include convolution operations, therefore requiring multiplications (see \#DSP blocks in \tabref{tab:perf_comparison}). Additionally, a clear improvement to the architecture presented in \cite{aicas2024} can be seen, since the current design requires fewer resources and still allows for faster inference. 

\bgroup
\def\arraystretch{1.2}
\begin{table}
	\centering
	\caption{Comparison to other hardware accelerator architectures for the MNIST dataset.}
	\begin{tabular}{|p{2.1cm}|p{1.5cm}|p{1.3cm}|p{1cm}|p{1cm}|}
	\hline
	Design                       & \cite{syncnn}        & \cite{fang}             & \cite{aicas2024}   & This work\\\hline
	FPGA / Board                 & ZCU102               & XCZU9EG                 & ZCU102             & ZCU102\\\hline
	Network\footnotemark[1]      & 32c5-p2-64c5-p2-2048 & 32c3-32c3-64c3-64c3-512 & 800-512-256        & 800-512-256\\\hline
	Clock freq. [MHz]            & 200                  & 125                     & \textbf{365}       & 300\\\hline
	\#LUTs                       & 224690               & 155952                  & 64156              & \textbf{56520}\\\hline
	\#block RAMs                 & Not reported         & 282                     & \textbf{91}        & \textbf{91}\\\hline
	\#DSP blocks                 & 555                  & 1795                    & \textbf{0}         & \textbf{0}\\\hline
	Images / second              & 1631                 & 2124                    & 2802               & \textbf{3400}\\\hline
	Energy [W]                   & Not reported         & 4.5                     & \textbf{2.097}     & 2.44\\\hline
	Accuracy [\%]                & \textbf{99.6}        & 99.2                    & 99.03\footnotemark[2] & 99.03\\\hline
	\end{tabular}
	\label{tab:perf_comparison}
\end{table}
\egroup
\footnotetext[1]{Number of neurons in last layer (10) deliberately omitted}
\footnotetext[2]{The previously presented accelerator architecture was synthesized again for the current network, which, per definition, must yield the same accuracy as the current design assuming the accelerator does not introduce any error}

\section{Conclusion}
This paper demonstrated that incorporating the encoding into the learning process of the network allows the network to learn which data points are most important and can thus positively influence inference accuracy. Furthermore, the paper introduced an \ac{snn} accelerator architecture for efficiently processing differential-time spikes, directly improving a previously introduced architecture. The new architecture requires fewer resources while providing faster inference on the tested MNIST dataset. Both contributions combined lead to an \ac{snn} architecture that achieves more than 99\% accuracy on the MNIST dataset without requiring any multiplications, while still being able to be implemented on medium-sized FPGAs. 

\bibliographystyle{IEEEtran}
\balance
\bibliography{bibliography.bib}

\begin{thebibliography}{10}
\providecommand{\url}[1]{#1}
\csname url@samestyle\endcsname
\providecommand{\newblock}{\relax}
\providecommand{\bibinfo}[2]{#2}
\providecommand{\BIBentrySTDinterwordspacing}{\spaceskip=0pt\relax}
\providecommand{\BIBentryALTinterwordstretchfactor}{4}
\providecommand{\BIBentryALTinterwordspacing}{\spaceskip=\fontdimen2\font plus
\BIBentryALTinterwordstretchfactor\fontdimen3\font minus
  \fontdimen4\font\relax}
\providecommand{\BIBforeignlanguage}[2]{{%
\expandafter\ifx\csname l@#1\endcsname\relax
\typeout{** WARNING: IEEEtran.bst: No hyphenation pattern has been}%
\typeout{** loaded for the language `#1'. Using the pattern for}%
\typeout{** the default language instead.}%
\else
\language=\csname l@#1\endcsname
\fi
#2}}
\providecommand{\BIBdecl}{\relax}
\BIBdecl

\bibitem{mnist}
\BIBentryALTinterwordspacing
Y.~LeCun, C.~Cortes, and C.~Burges, ``{MNIST Handwritten Digit Database},''
  2010. [Online]. Available: \url{http://yann.lecun.com/exdb/mnist}
\BIBentrySTDinterwordspacing

\bibitem{shd_dataset}
\BIBentryALTinterwordspacing
B.~Cramer, Y.~Stradmann, J.~Schemmel, and F.~Zenke, ``The heidelberg spiking
  datasets for the systematic evaluation of spiking neural networks,''
  \emph{CoRR}, vol. abs/1910.07407, 2019. [Online]. Available:
  \url{http://arxiv.org/abs/1910.07407}
\BIBentrySTDinterwordspacing

\bibitem{tf_speechcommands}
\BIBentryALTinterwordspacing
P.~{Warden}, ``{Speech Commands: A Dataset for Limited-Vocabulary Speech
  Recognition},'' \emph{ArXiv e-prints}, Apr. 2018. [Online]. Available:
  \url{https://arxiv.org/abs/1804.03209}
\BIBentrySTDinterwordspacing

\bibitem{darwin}
D.~Ma, J.~Shen, Z.~Gu, M.~Zhang, X.~Zhu, X.~Xu, Q.~Xu, Y.~Shen, and G.~Pan,
  ``Darwin: A neuromorphic hardware co-processor based on spiking neural
  networks,'' \emph{Journal of Systems Architecture}, vol.~77, pp. 43--51,
  2017.

\bibitem{energy_efficient_neuromorph_arch}
\BIBentryALTinterwordspacing
Q.~Wang, Y.~Li, B.~Shao, S.~Dey, and P.~Li, ``Energy efficient parallel
  neuromorphic architectures with approximate arithmetic on fpga,''
  \emph{Neurocomputing}, vol. 221, pp. 146--158, 2017. [Online]. Available:
  \url{https://www.sciencedirect.com/science/article/pii/S0925231216311213}
\BIBentrySTDinterwordspacing

\bibitem{loihi}
M.~Davies, A.~Wild, G.~Orchard, Y.~Sandamirskaya, G.~A.~F. Guerra, P.~Joshi,
  P.~Plank, and S.~R. Risbud, ``Advancing neuromorphic computing with loihi: A
  survey of results and outlook,'' \emph{Proceedings of the IEEE}, vol. 109,
  no.~5, pp. 911--934, 2021.

\bibitem{noc}
H.~Fang, A.~Shrestha, D.~Ma, and Q.~Qiu, ``Scalable noc-based neuromorphic
  hardware learning and inference,'' in \emph{IJCNN}, 2018, pp. 1--8.

\bibitem{snava}
A.~Sripad, G.~Sanchez, M.~Zapata, V.~Pirrone, T.~Dorta, S.~Cambria, A.~Marti,
  K.~Krishnamourthy, and J.~Madrenas, ``Snava—a real-time multi-fpga
  multi-model spiking neural network simulation architecture,'' \emph{Neural
  Networks}, vol.~97, pp. 28--45, 2018.

\bibitem{TCASSPIKEMNIST21}
S.~Li, Z.~Zhang, R.~Mao, J.~Xiao, L.~Chang, and J.~Zhou, ``{A Fast and
  Energy-Efficient SNN Processor With Adaptive Clock/Event-Driven Computation
  Scheme and Online Learning},'' \emph{IEEE Trans. on Circuits and Systems I:
  Regular Papers}, vol.~68, no.~4, pp. 1543--1552, 2021.

\bibitem{truenorth}
F.~Akopyan, J.~Sawada, A.~Cassidy, R.~Alvarez-Icaza, J.~Arthur, P.~Merolla,
  N.~Imam, Y.~Nakamura, P.~Datta, G.-J. Nam, B.~Taba, M.~Beakes, B.~Brezzo,
  J.~B. Kuang, R.~Manohar, W.~P. Risk, B.~Jackson, and D.~S. Modha,
  ``Truenorth: Design and tool flow of a 65 mw 1 million neuron programmable
  neurosynaptic chip,'' \emph{IEEE Trans. on Computer-Aided Design of
  Integrated Circuits and Systems}, vol.~34, no.~10, pp. 1537--1557, 2015.

\bibitem{low_power_fpga_snn_acc}
H.~Liu, Y.~Chen, Z.~Zeng, M.~Zhang, and H.~Qu, ``A low power and low latency
  fpga-based spiking neural network accelerator,'' in \emph{2023 International
  Joint Conference on Neural Networks (IJCNN)}, 2023, pp. 1--8.

\bibitem{spinnaker}
S.~B. Furber, F.~Galluppi, S.~Temple, and L.~A. Plana, ``The spinnaker
  project,'' \emph{Proceedings of the IEEE}, vol. 102, no.~5, pp. 652--665,
  2014.

\bibitem{aicas2024}
\BIBentryALTinterwordspacing
D.~Windhager, B.~A. Moser, and M.~Lunglmayr, ``Snn architecture for
  differential time encoding using decoupled processing time,'' 2023. [Online].
  Available: \url{https://arxiv.org/abs/2311.14447}
\BIBentrySTDinterwordspacing

\bibitem{transformers}
\BIBentryALTinterwordspacing
A.~Vaswani, N.~Shazeer, N.~Parmar, J.~Uszkoreit, L.~Jones, A.~N. Gomez,
  L.~Kaiser, and I.~Polosukhin, ``Attention is all you need,'' \emph{CoRR},
  vol. abs/1706.03762, 2017. [Online]. Available:
  \url{http://arxiv.org/abs/1706.03762}
\BIBentrySTDinterwordspacing

\bibitem{odesa}
\BIBentryALTinterwordspacing
A.~Mehrabi, Y.~Bethi, A.~{van Schaik}, and S.~Afshar, ``An optimized
  multi-layer spiking neural network implementation in fpga without
  multipliers,'' \emph{Procedia Computer Science}, vol. 222, pp. 407--414,
  2023, international Neural Network Society Workshop on Deep Learning
  Innovations and Applications (INNS DLIA 2023). [Online]. Available:
  \url{https://www.sciencedirect.com/science/article/pii/S1877050923009444}
\BIBentrySTDinterwordspacing

\bibitem{spiker}
\BIBentryALTinterwordspacing
A.~Carpegna, A.~Savino, and S.~D. Carlo, ``Fpga-optimized hardware acceleration
  for spiking neural networks,'' \emph{CoRR}, vol. abs/2201.06993, 2022.
  [Online]. Available: \url{https://arxiv.org/abs/2201.06993}
\BIBentrySTDinterwordspacing

\bibitem{spiker+}
\BIBentryALTinterwordspacing
------, ``Spiker+: a framework for the generation of efficient spiking neural
  networks fpga accelerators for inference at the edge,'' 2024. [Online].
  Available: \url{https://arxiv.org/abs/2401.01141}
\BIBentrySTDinterwordspacing

\bibitem{minitaur}
D.~Neil and S.-C. Liu, ``Minitaur, an event-driven fpga-based spiking network
  accelerator,'' \emph{IEEE Transactions on Very Large Scale Integration (VLSI)
  Systems}, vol.~22, no.~12, pp. 2621--2628, 2014.

\bibitem{fast_lp_lod}
\BIBentryALTinterwordspacing
M.~S. Ansari, S.~Gandhi, B.~F. Cockburn, and J.~Han, ``Fast and low-power
  leading-one detectors for energy-efficient logarithmic computing,'' \emph{IET
  Computers \& Digital Techniques}, vol.~15, no.~4, pp. 241--250, 2021.
  [Online]. Available:
  \url{https://ietresearch.onlinelibrary.wiley.com/doi/abs/10.1049/cdt2.12019}
\BIBentrySTDinterwordspacing

\bibitem{vlsi_lod}
K.~Kunaraj and R.~Seshasayanan, ``Leading one detectors and leading one
  position detectors - an evolutionary design methodology,'' \emph{Canadian
  Journal of Electrical and Computer Engineering}, vol.~36, no.~3, pp.
  103--110, 2013.

\bibitem{vlsi_lp_lod}
K.~Abed and R.~Siferd, ``Vlsi implementations of low-power leading-one detector
  circuits,'' in \emph{Proceedings of the IEEE SoutheastCon 2006}, 2006, pp.
  279--284.

\bibitem{approx_lod}
S.~Gandhi, M.~S. Ansari, B.~F. Cockburn, and J.~Han, ``Approximate leading one
  detector design for a hardware-efficient mitchell multiplier,'' in \emph{2019
  IEEE Canadian Conference of Electrical and Computer Engineering (CCECE)},
  2019, pp. 1--4.

\bibitem{snntorch}
J.~K. Eshraghian, M.~Ward, E.~Neftci, X.~Wang, G.~Lenz, G.~Dwivedi,
  M.~Bennamoun, D.~S. Jeong, and W.~D. Lu, ``Training spiking neural networks
  using lessons from deep learning,'' \emph{Proceedings of the IEEE}, vol. 111,
  no.~9, pp. 1016--1054, 2023.

\bibitem{syncnn}
S.~Panchapakesan, Z.~Fang, and J.~Li, ``Syncnn: Evaluating and accelerating
  spiking neural networks on fpgas,'' in \emph{2021 31st International
  Conference on Field-Programmable Logic and Applications (FPL)}, 2021, pp.
  286--293.

\bibitem{fang}
H.~Fang, Z.~Mei, A.~Shrestha, Z.~Zhao, Y.~Li, and Q.~Qiu, ``Encoding, model,
  and architecture: Systematic optimization for spiking neural network in
  fpgas,'' in \emph{2020 IEEE/ACM International Conference On Computer Aided
  Design (ICCAD)}, 2020, pp. 1--9.

\end{thebibliography}

\end{document}